\documentclass[11pt]{article}
\usepackage[hyperref]{acl}
\usepackage{times}
\usepackage{latexsym}
\usepackage{graphicx}
\usepackage{booktabs}
\usepackage{amsmath}
\usepackage{multirow}
\usepackage[T1]{fontenc}
\usepackage[utf8]{inputenc}
\usepackage{microtype}
\usepackage{inconsolata}
\usepackage{url}
\usepackage{hyperref}
\usepackage{tikz}
\usetikzlibrary{arrows.meta,positioning}

\title{Domain-Specific Hallucination Detection in Large Language Models}

\author{Varun Teja Chundru \and Debasmita Biswas \\
  Department of Computer Science \\
  Purdue University Fort Wayne \\
  \texttt{\{vchundru, biswd01\}@pfw.edu}}

\begin{document}
\maketitle

\begin{abstract}
Large language models generate fluent text that can contain unfaithful claims---a
phenomenon known as hallucination. We present a multi-signal detection pipeline
combining fine-tuned DeBERTa-v3 classification, Monte Carlo (MC) Dropout uncertainty
quantification, and temperature-scaled calibration for response-level hallucination
detection. Evaluated on the HaluEval benchmark, our pipeline achieves F1=0.915 and
AUROC=0.977 on general-domain tasks, with per-task F1 scores of 0.97 (QA), 0.96
(Summarization), and 0.82 (Dialogue). MC Dropout inference further improves accuracy
to 93.2\%. A context ablation study confirms the model performs genuine entailment
reasoning rather than exploiting surface patterns, with summarization F1 dropping
24\% when knowledge context is removed. Learning curve analysis reveals that 25\% of
training data captures 77\% of full-data performance. Beyond detection, we apply
Direct Preference Optimization (DPO) to a Qwen2.5-0.5B generator, reducing its
hallucination rate from 85.5\% to 37.7\% (55.9\% relative reduction) as measured by
our detector. Cross-domain evaluation on the SciFact biomedical benchmark shows that
general-domain training transfers poorly (F1=0.52), motivating domain-specific
fine-tuning. PubMedBERT fine-tuned on SciFact achieves F1=0.63 and AUROC=0.81,
demonstrating that domain-matched pre-training is the strongest adaptation strategy.
Code and models are available at \url{https://github.com/varunteja99/hallucination-detection-nlp}.
\end{abstract}

\section{Introduction}

Large language models (LLMs) produce text that is syntactically fluent and contextually
plausible but can contain fabricated facts, misattributed claims, and unsupported
inferences \citep{ji2023survey}. These hallucinations pose a significant barrier to
deploying LLMs in high-stakes domains such as medicine, law, and scientific research,
where factual accuracy is essential.

The hallucination detection problem can be formulated as a natural language inference
(NLI) task: given a knowledge source $K$, a prompt $Q$, and a generated response $R$,
determine whether $R$ is faithful to or hallucinated with respect to $K$. While prior
work has explored entailment-based and retrieval-based approaches, two critical gaps
remain. First, single-model detectors provide point estimates without conveying
prediction confidence, leaving practitioners unable to distinguish high-certainty
detections from ambiguous cases. Second, detectors trained on general-domain
benchmarks often fail on specialized domains where terminology and reasoning patterns
differ substantially.

This paper makes four contributions, organized around three experiments---detection,
mitigation, and cross-domain transfer. (1) We develop a multi-signal detection
pipeline combining fine-tuned DeBERTa-v3 with MC Dropout uncertainty,
achieving F1=0.915 and AUROC=0.977 on HaluEval. (2) We conduct ablation
studies---context removal, learning curves, and ensemble analysis---characterizing
when and why the detector succeeds. (3) We show that DPO reduces hallucination
rates by 55.9\% in a generator, evaluated by our detector in a closed-loop setup.
(4) We evaluate cross-domain transfer to SciFact and show that PubMedBERT
fine-tuning achieves AUROC=0.808, demonstrating that domain-matched pre-training
is the most effective adaptation strategy.

\section{Related Work}

\paragraph{Hallucination Detection.} Hallucination in LLMs has been categorized into
intrinsic hallucination (contradicting the source) and extrinsic hallucination
(introducing unverifiable claims) \citep{ji2023survey}. Detection approaches span
entailment-based classification \citep{honovich2022true}, retrieval-augmented
verification \citep{min2023factscore}, and uncertainty estimation
\citep{kuhn2023semantic}. \citet{li2023halueval} introduced the HaluEval benchmark
with task-specific hallucinated samples generated via ChatGPT, providing a controlled
evaluation framework across QA, dialogue, and summarization. Our work builds on this
benchmark while extending the analysis with uncertainty quantification and
cross-domain evaluation.

\paragraph{Uncertainty Quantification.} Monte Carlo Dropout
\citep{gal2016dropout} provides a practical approximation to Bayesian inference by
performing multiple stochastic forward passes with dropout enabled at test time. The
variance across passes captures epistemic uncertainty, which has been applied to
out-of-distribution detection \citep{lakshminarayanan2017simple} and selective
prediction. We integrate MC Dropout into our detection pipeline, showing it improves
accuracy from 91.3\% to 93.2\%.

\paragraph{Preference Optimization.} Direct Preference Optimization (DPO) frames
alignment as a classification problem over preference pairs, avoiding the instability
of reinforcement learning from human feedback \citep{rafailov2023direct}. While DPO has
primarily been applied to safety and helpfulness alignment, we apply it specifically
to hallucination reduction, using faithful and hallucinated responses as preference
pairs.

\section{Methodology}

\subsection{Detection Pipeline}

\begin{figure}[t]
\centering
\scriptsize
\begin{tikzpicture}[
  node distance=1.5mm and 2mm,
  every node/.style={align=center,inner sep=2pt,font=\scriptsize},
  blk/.style={draw,rounded corners=1.5pt,fill=blue!6,minimum height=4mm},
  sig/.style={draw,rounded corners=1.5pt,fill=green!9,minimum height=4mm},
  ens/.style={draw,rounded corners=1.5pt,fill=orange!12,minimum height=4mm},
  ar/.style={-Stealth,thin}
]
\node[blk] (in) {$[Q;K;R]$};
\node[blk,right=of in] (deb) {DeBERTa-v3};
\node[sig,right=4mm of deb,yshift=5.5mm] (s1) {Single: $P_{\text{std}}$};
\node[sig,right=4mm of deb] (s2) {MC: $\bar p,\sigma$};
\node[sig,right=4mm of deb,yshift=-5.5mm] (s3) {Cal: $P_{\text{cal}}$};
\node[ens,right=4mm of s2] (e) {Ensembles};
\draw[ar] (in) -- (deb);
\draw[ar] (deb.east) -- (s1.west);
\draw[ar] (deb.east) -- (s2.west);
\draw[ar] (deb.east) -- (s3.west);
\draw[ar] (s1.east) -- (e.west);
\draw[ar] (s2.east) -- (e.west);
\draw[ar] (s3.east) -- (e.west);
\end{tikzpicture}
\caption{Detection pipeline. DeBERTa-v3 produces three signals---single pass, MC
Dropout ($T{=}20$), and temperature-scaled---combined by Simple Average or LR
Meta-Classifier.}
\label{fig:pipeline}
\end{figure}
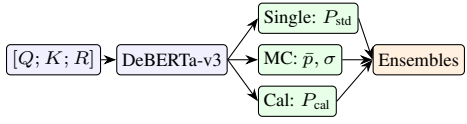

Our detection pipeline (Figure~\ref{fig:pipeline}) builds three inference modes
on a shared fine-tuned DeBERTa-v3 backbone, plus two ensembles.

\paragraph{DeBERTa-v3 Classifier.} We use DeBERTa-v3-base \citep{he2023debertav3} as
our core classifier. DeBERTa employs a disentangled attention mechanism that
separates content and position representations into distinct vectors, computing
attention weights using disentangled matrices for content-to-content,
content-to-position, and position-to-content interactions. An enhanced mask decoder
aggregates these signals to produce context-aware token representations. We add a
classification head and fine-tune for binary NLI: given a concatenated input
$[Q; K; R]$, the model outputs $P(\text{hallucinated} \mid Q, K, R)$. Training uses
AdamW with learning rate $2 \times 10^{-5}$, linear warmup over 10\% of steps, batch
size 16, and 3 epochs with fp32 mixed precision.

\paragraph{MC Dropout Uncertainty.} A single forward pass yields an overconfident
point estimate---small logit shifts produce large probability changes near the
decision boundary. We instead keep dropout active at inference and run $T = 20$
stochastic forward passes. The mean
$\bar p = \tfrac{1}{T}\sum_{t} p_t$ serves as the prediction and the standard
deviation $\sigma = \sqrt{\tfrac{1}{T}\sum_{t}(p_t - \bar p)^2}$ captures
epistemic uncertainty. Averaging across stochastic sub-networks reduces variance
and smooths probabilities near the boundary, both improving top-1 accuracy and
producing a $\sigma$ signal that is high precisely on ambiguous inputs.

\paragraph{Temperature Scaling.} We learn a scalar $T^*$ on the validation set by
minimizing NLL: $T^* = \arg\min_T L_{\text{NLL}}(\text{softmax}(z/T), y)$.
Dividing logits by $T^*$ before softmax sharpens or flattens the distribution
without changing the argmax, so accuracy and F1 are unchanged but probabilities
become better calibrated. This matters because uncalibrated MC Dropout variance is
suppressed on uncertain examples and NLL is inflated on wrong-but-confident
predictions, causing threshold instability across domains.

\paragraph{Ensemble Methods.} We evaluate two ensembles. \emph{Simple Average} takes
the unweighted mean of $P_{\text{std}}$ and $\bar p$, the single-pass and MC Dropout
mean probabilities. \emph{LR Meta-Classifier} is a logistic regression trained on the
validation set whose four features are $P_{\text{std}}$, $\bar p$, $\sigma$, and the
cosine similarity between MiniLM embeddings of context and response; it learns
optimal feature weights rather than assuming equal contribution.

\subsection{DPO Hallucination Mitigation}

Beyond detection, we train a generator model to produce fewer hallucinations using
DPO \citep{rafailov2023direct}. We construct preference pairs from HaluEval: for each
prompt, the reference (faithful) answer is the chosen response and the hallucinated
answer is the rejected response. We fine-tune Qwen2.5-0.5B-Instruct
\citep{qwen2025qwen25} on 21K preference pairs for 1 epoch with learning rate
$5 \times 10^{-6}$ and $\beta = 0.1$. At evaluation, our DeBERTa detector scores
held-out generations from both the base and DPO-trained models, providing a
detector-in-the-loop assessment of hallucination reduction.

\subsection{Cross-Domain Adaptation}

For SciFact, we evaluate three adaptation strategies trading off pre-training
corpus, NLI priors, and target-domain training: (A) fine-tuning DeBERTa-v3 on
SciFact (in-domain training without domain pre-training); (B) fine-tuning
PubMedBERT \citep{gu2021domain}, pre-trained on PubMed abstracts and PMC
full-text, on SciFact (domain-matched pre-training); and (C) sequential
transfer---DeBERTa-v3 fine-tuned first on HaluEval, then on SciFact---combining
NLI priors with target-domain adaptation.

\section{Experimental Setup}

\subsection{Datasets}

\paragraph{HaluEval.} The HaluEval benchmark \citep{li2023halueval} contains 30,000
samples across three tasks---QA, Dialogue, and Summarization---each with 10,000
balanced examples (5,000 faithful, 5,000 hallucinated). Hallucinated responses were
generated by ChatGPT with task-specific prompting. We use a 70/15/15 stratified split
(21,000 train / 4,500 validation / 4,500 test) with a fixed random seed for
reproducibility.

\paragraph{SciFact.} SciFact \citep{wadden2020fact} is a biomedical claim
verification dataset containing 1,109 scientific claims paired with evidence from a
corpus of 5,183 abstracts. We extract 693 labeled (claim, evidence) pairs and apply a
stratified 70/15/15 split (484 train / 103 validation / 106 test). Labels are
binarized: SUPPORT $\rightarrow$ faithful (0), CONTRADICT $\rightarrow$ hallucinated (1).

\subsection{Baselines}

We compare our fine-tuned detector against two baselines that isolate the
contribution of training and uncertainty quantification respectively. (1)
\textbf{Zero-shot DeBERTa-v3-MNLI:} the pre-trained model without HaluEval
fine-tuning, evaluating off-the-shelf NLI transfer. (2) \textbf{Standard inference:}
single-pass fine-tuned DeBERTa without MC Dropout, calibration, or ensembling. The
context ablation in Section~\ref{sec:context-ablation} is reported separately as a
diagnostic study, not as a competing baseline.

\subsection{Metrics}

We report Accuracy, F1-score, and AUROC. Accuracy measures overall classification
correctness. F1-score is the harmonic mean of precision and recall, important
because hallucinated samples in HaluEval are balanced but real-world distributions
are skewed. AUROC (Area Under the Receiver Operating Characteristic Curve) evaluates
ranking quality across all thresholds, capturing how well the model separates
classes independently of a fixed decision boundary.

\section{Experiment 1: Hallucination Detection}

\subsection{Main Detection Results}

Table~\ref{tab:main} reports the six rows of our detection comparison on the
HaluEval test set. \textbf{Zero-shot DeBERTa} is the off-the-shelf MNLI model with
no HaluEval fine-tuning, establishing a transfer baseline. \textbf{Fine-tuned
DeBERTa} is the same backbone after 3 epochs of fine-tuning on HaluEval, evaluated
with a single deterministic forward pass. \textbf{MC Dropout mean} uses the same
fine-tuned weights but enables dropout at inference and averages probabilities over
20 stochastic passes, smoothing the decision boundary. \textbf{Calibrated DeBERTa}
applies the learned temperature $T^* {=} 1.69$ to the fine-tuned logits before
softmax; because temperature scaling preserves argmax, accuracy and F1 are
identical to the fine-tuned row by construction---the value lies in better-calibrated
probabilities for downstream uncertainty use. \textbf{Simple Average} takes the
unweighted mean of $P_{\text{std}}$ and $\bar p$. \textbf{LR Meta-Classifier} trains
a logistic regression on the validation set with four features: $P_{\text{std}}$,
$\bar p$, $\sigma$, and retrieval similarity.

\begin{table}[h]
\centering
\small
\begin{tabular}{lccc}
\toprule
\textbf{Method} & \textbf{Acc} & \textbf{F1} & \textbf{AUROC} \\
\midrule
Zero-shot DeBERTa     & 0.500 & 0.430 & 0.650 \\
Fine-tuned DeBERTa    & 0.913 & 0.915 & 0.977 \\
MC Dropout mean       & \textbf{0.932} & \textbf{0.931} & 0.978 \\
Calibrated DeBERTa    & 0.913 & 0.915 & 0.977 \\
Simple Average        & 0.920 & 0.921 & \textbf{0.979} \\
LR Meta-Classifier    & 0.931 & 0.930 & 0.960 \\
\bottomrule
\end{tabular}
\caption{Detection performance on HaluEval test set (4{,}500 samples). MC Dropout
provides the best single-model accuracy and F1 by averaging 20 stochastic forward
passes; Simple Average gives the highest AUROC by averaging the two DeBERTa-based
probability estimates.}
\label{tab:main}
\end{table}

The headline result is that MC Dropout improves accuracy from 91.3\% to 93.2\%
(+1.9 points) and F1 from 0.915 to 0.931 (+0.016) over single-pass inference, with
no additional training. This confirms that the variance-reduction effect of
averaging stochastic sub-networks meaningfully improves the detector on cases where
a single pass would land on the wrong side of the decision boundary. AUROC moves
only marginally (0.977 $\rightarrow$ 0.978) because ranking quality already
saturates with the fine-tuned model; the gain is concentrated near threshold.

The LR Meta-Classifier matches MC Dropout on accuracy (0.931) but loses AUROC
(0.960 vs. 0.978). The cause is the retrieval similarity feature: cosine
similarity between MiniLM embeddings of context and response achieves only
AUROC $\approx$ 0.38 in isolation---faithful and hallucinated responses share
surface vocabulary in HaluEval, so this near-random feature degrades ranking
quality even after the LR weights it down.

\subsection{Per-Task Analysis}

Table~\ref{tab:per-task} breaks down fine-tuned DeBERTa performance by HaluEval
subtask. QA is easiest (F1=0.97) due to strong lexical overlap between questions
and factoid answers---hallucinated answers typically substitute incorrect entities
or numbers detectable from the question alone. Summarization is also strong
(F1=0.96) given the source document. Dialogue is hardest (F1=0.82) because
conversational responses are shorter, more implicit, and contain fewer lexical
anchors to the knowledge source.

\begin{table}[h]
\centering
\small
\begin{tabular}{lccc}
\toprule
\textbf{Task} & \textbf{Acc} & \textbf{F1} & \textbf{AUROC} \\
\midrule
QA              & 0.967 & 0.970 & 0.996 \\
Summarization   & 0.940 & 0.960 & 0.988 \\
Dialogue        & 0.830 & 0.820 & 0.944 \\
\midrule
Overall         & 0.913 & 0.915 & 0.977 \\
\bottomrule
\end{tabular}
\caption{Per-task detection performance of fine-tuned DeBERTa on HaluEval.}
\label{tab:per-task}
\end{table}

\section{Detector Analysis}
\label{sec:analysis}

\subsection{Context Ablation}
\label{sec:context-ablation}

\begin{figure}[h]
\centering
\includegraphics[width=0.95\columnwidth]{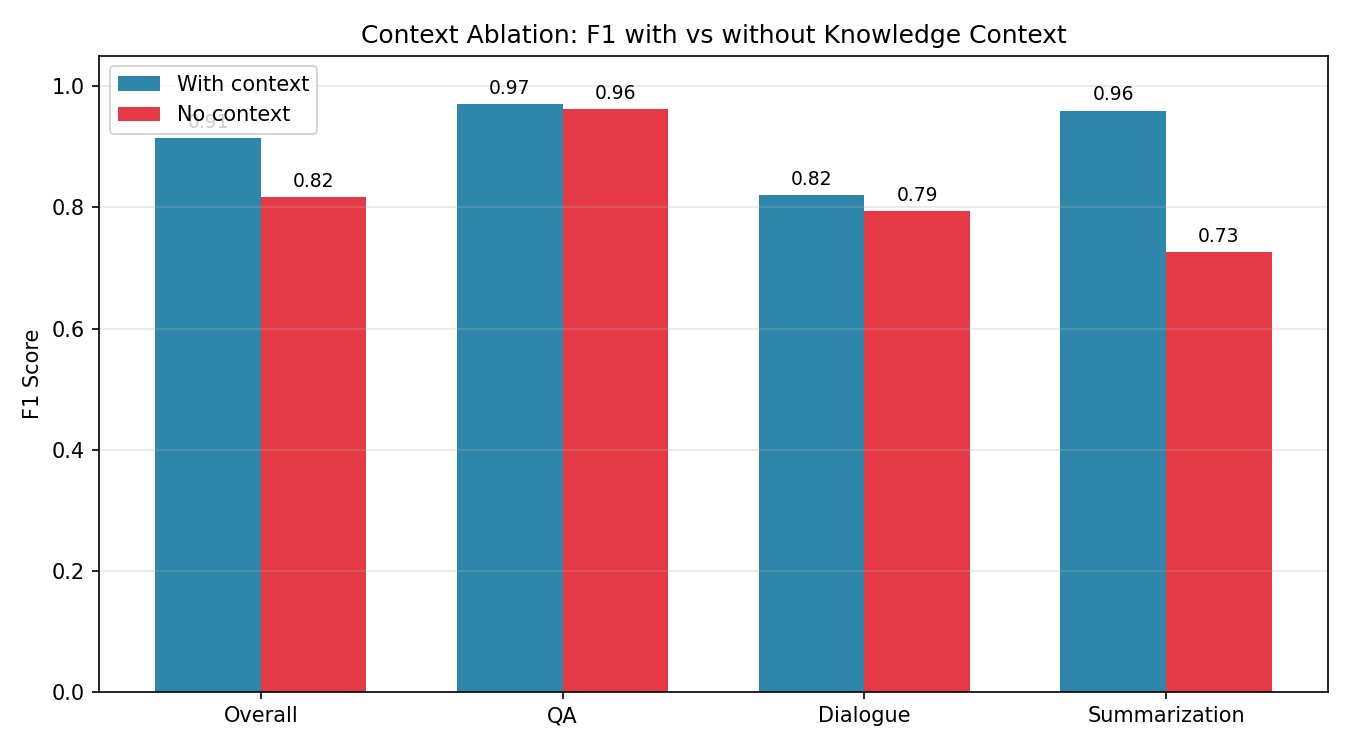}
\caption{F1 with and without knowledge context across tasks. Summarization depends
most on context ($-$24\%); QA is largely self-contained ($-$1\%).}
\label{fig:ctx-ablation}
\end{figure}

To determine whether the model performs genuine entailment reasoning or exploits
surface-level shortcuts in the response alone, we strip the knowledge context $K$
from all test inputs (keeping only $[Q;R]$) and re-evaluate the same fine-tuned
weights. Figure~\ref{fig:ctx-ablation} shows the result.

Overall F1 drops from 0.91 to 0.82 without context, confirming that the model
leverages the knowledge source rather than memorizing surface artifacts. The effect
is task-dependent. Summarization F1 drops 24\% (0.96 $\rightarrow$ 0.73), indicating
that detecting hallucinated summaries requires comparing against the source
document---unsurprising, since a summary's faithfulness is by definition relative
to its source. QA F1 drops only 1\% (0.97 $\rightarrow$ 0.96), suggesting that
factoid QA hallucinations are often detectable from the question--answer pair alone,
likely because hallucinated answers contain implausible entity substitutions or
numerical inconsistencies the model can flag without re-reading the passage.
Dialogue sits in between (0.82 $\rightarrow$ 0.79).

\subsection{Learning Curves}

\begin{figure}[h]
\centering
\includegraphics[width=0.95\columnwidth]{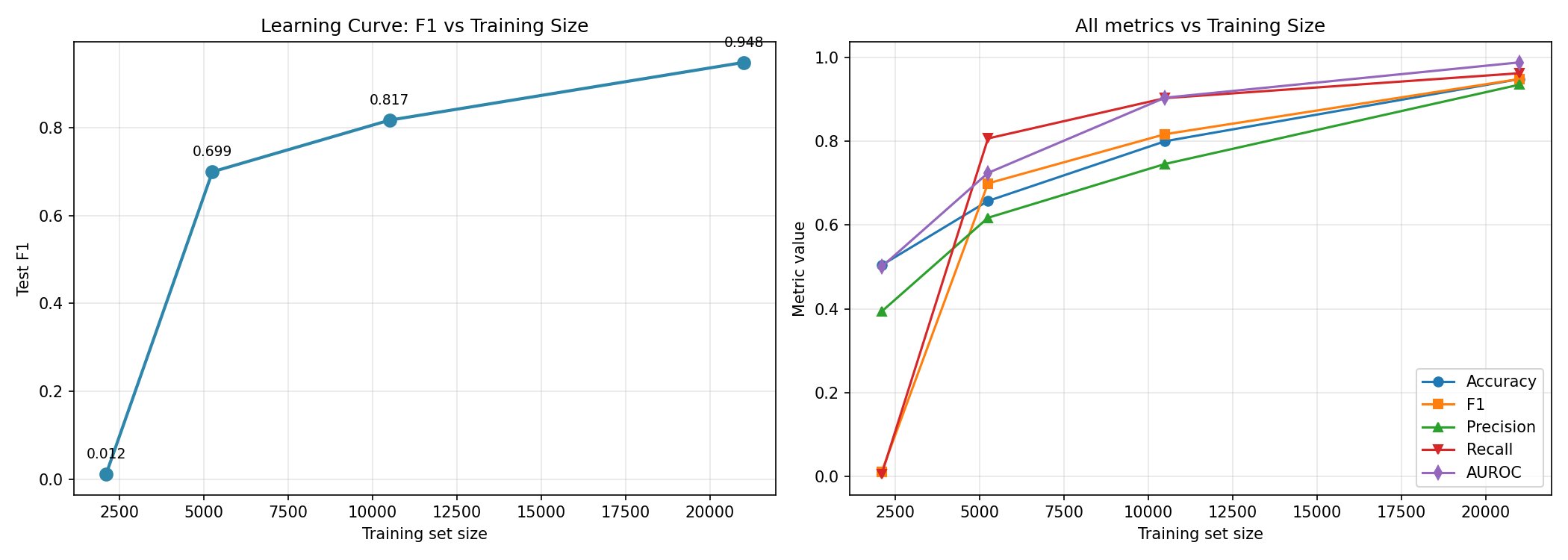}
\caption{Learning curves: F1 (left) and all metrics (right) versus training set
size. A sharp elbow at $\sim$5K examples captures most of the discriminative signal.}
\label{fig:learning}
\end{figure}

We re-train DeBERTa from scratch on 10\%, 25\%, 50\%, and 100\% of the HaluEval
training data to characterize data efficiency (Figure~\ref{fig:learning}).

At 10\% (2.1K examples), the model fails completely (F1=0.01), unable to
distinguish the classes. At 25\% (5.3K), F1 jumps to 0.70---a sharp elbow
indicating that approximately 5K labeled examples are sufficient to learn the core
discrimination signal. Performance continues improving to 0.82 at 50\% and 0.95 at
100\%, but with diminishing returns. This is practically important for new domains:
bootstrapping a usable detector requires only $\sim$5K labeled examples, not the
full 21K we used.

\section{Experiment 2: Mitigation via DPO}

The detector is useful in itself, but a stronger test of its utility is whether it
can drive a generator to produce fewer hallucinations. Table~\ref{tab:dpo}
summarizes this closed-loop experiment.

\begin{table}[h]
\centering
\small
\begin{tabular}{lcc}
\toprule
\textbf{Model} & \textbf{Hall.\ Rate} & \textbf{Mean $P$(hall)} \\
\midrule
Base (Qwen2.5-0.5B)  & 0.855  & 0.816 \\
DPO-trained          & 0.377  & 0.293 \\
\midrule
Absolute reduction   & $-0.477$ & $-0.523$ \\
Relative reduction   & 55.9\%   & 64.1\% \\
\bottomrule
\end{tabular}
\caption{DPO hallucination reduction on 4,500 held-out test generations. Our
DeBERTa detector evaluates both base and DPO generations.}
\label{tab:dpo}
\end{table}

The base Qwen2.5-0.5B-Instruct model produces hallucinated responses for 85.5\% of
held-out test prompts, as scored by our DeBERTa detector. After DPO training on 21K
preference pairs, the hallucination rate drops to 37.7\%---a 55.9\% relative
reduction. The probability distribution shifts substantially: the base model
concentrates near $P(\text{hall}) = 1.0$ while the DPO model shifts mass toward
$P(\text{hall}) = 0$, with mean detector probability falling from 0.816 to 0.293.

Two caveats are worth stating. First, the detector and the DPO preference signal
share supervision (both derive from HaluEval pairs), so the 55.9\% number is a
co-evaluation rather than a fully held-out test. Second, the detector serves here
as a consistent automated evaluation signal in the detector-in-the-loop paradigm
rather than a gold-standard verdict.

\section{Experiment 3: Cross-Domain Transfer}

Applying the HaluEval-trained DeBERTa zero-shot to SciFact biomedical claims yields
F1=0.517 and AUROC=0.515---barely above chance. The model predicts nearly all
scientific claims as hallucinated because biomedical claim--evidence pairs differ
substantially from HaluEval's ChatGPT-generated responses: scientific claims use
technical vocabulary, hedged language, and citation-grounded reasoning the
source-domain training never saw.

To address this, we evaluate the three adaptation configurations from
Section 3.3 on 484 SciFact training examples.

\begin{table}[h]
\centering
\small
\setlength{\tabcolsep}{4pt}
\begin{tabular}{llccc}
\toprule
\textbf{Cfg} & \textbf{Base Model} & \textbf{Acc} & \textbf{F1} & \textbf{AUROC} \\
\midrule
Zero & HaluEval-DeBERTa  & 0.349 & 0.517 & 0.515 \\
A    & DeBERTa-v3        & 0.604 & 0.488 & 0.582 \\
B    & PubMedBERT        & \textbf{0.764} & \textbf{0.627} & \textbf{0.808} \\
C    & HaluEval$\rightarrow$SciFact & 0.472 & 0.533 & 0.610 \\
\bottomrule
\end{tabular}
\caption{Cross-domain results on SciFact (106 test examples). Config B
(PubMedBERT \citep{gu2021domain} fine-tuned on SciFact) wins on every metric.}
\label{tab:scifact}
\end{table}

PubMedBERT (Config B) achieves the strongest results with F1=0.627 and AUROC=0.808,
demonstrating that domain-matched pre-training provides the largest benefit for
biomedical claim verification. Config C (HaluEval$\rightarrow$SciFact transfer)
outperforms zero-shot on AUROC (0.610 vs. 0.515), indicating that general-domain NLI
pre-training provides useful initialization. Config A (DeBERTa fine-tuned on SciFact
alone) achieves higher accuracy than zero-shot (0.604 vs. 0.349) but lower F1
(0.488 vs. 0.517) because it learns a more conservative threshold but lacks both
the domain vocabulary of PubMedBERT and the NLI priors from HaluEval. All
configurations remain below 0.7 F1 with only 484 training examples. The ranking
domain-matched pre-training $>$ source-task transfer $>$ in-domain training alone
$>$ zero-shot suggests that the dominant signal in cross-domain hallucination
detection is the pre-training corpus, not the fine-tuning data.

\section{Conclusion}

Our multi-signal hallucination detection pipeline achieves F1=0.915 on HaluEval,
with MC Dropout improving accuracy to 93.2\%. Diagnostic studies showed the
detector leverages knowledge context (overall F1 drops 10 points without it;
summarization most affected at 24\%) and that $\sim$5K labeled examples suffice
for usable performance. DPO training reduced generator hallucination rates by
55.9\% under our detector. On SciFact, PubMedBERT fine-tuning achieved
AUROC=0.808---ahead of source-task transfer and in-domain training alone.
Future work includes span-level localization, scaling DPO to larger generators,
and adapting to legal and financial text.

\bibliography{custom}

\end{document}